\documentclass[lettersize,journal]{IEEEtran}
\usepackage{amsfonts}
\usepackage[ruled,vlined]{algorithm2e}
\usepackage{array}
\usepackage[caption=false,font=normalsize,labelfont=sf,textfont=sf]{subfig}
\usepackage{textcomp}
\usepackage{stfloats}
\usepackage{url}
\usepackage{verbatim}
\usepackage{graphicx}
\usepackage{cite}
\usepackage{bm}
\usepackage{setspace}

\usepackage{amssymb}
\usepackage{amsthm,amsmath}
\usepackage{mathrsfs}
\usepackage{indentfirst}
\usepackage{multirow}
\usepackage[switch]{lineno}
\usepackage{algpseudocode}
\usepackage{algorithmicx}
\usepackage{algpseudocode}
\begin{document}

\title{Collaborative Position Reasoning Network for Referring Image Segmentation}
 
\author{Jianjian Cao, Beiya Dai, Yulin Li, Xiameng Qin and Jingdong Wang,~\IEEEmembership{Fellow,~IEEE}
        % <-this % stops a space
\thanks{Jianjian~Cao is with School of Information Science and Technology, Fudan University, Shanghai, China. (Email: jjcao22@m.fudan.edu.cn).}
\thanks{Beiya Dai is with Department of Computer Science, Notional University of Defense Technology, Changsha, China. (Email: beiya\_dai@nudt.edu.cn).}
\thanks{Yulin Li, Xiameng Qin and Jingdong Wang are with Baidu Inc., Beijing, China. (Email: liyulin03@baidu.com, qinxiameng@baidu.com, wangjingdong@baidu.com).}}

%\markboth{IEEE TRANSACTIONS ON IMAGE PROCESSING}%


\maketitle
\begin{abstract}
Given an image and a natural language expression as input, the goal of referring image segmentation is to segment the foreground masks of the entities referred by the expression. 
Existing methods mainly focus on interactive learning between vision and language to enhance the multi-modal representations for global context reasoning. However, predicting directly in pixel-level space can lead to collapsed positioning and poor segmentation results.
Its main challenge lies in how to explicitly model entity localization, especially for non-salient entities. 
In this paper, we tackle this problem by executing a Collaborative Position Reasoning Network (CPRN) via the proposed novel Row-and-Column interactive (RoCo) and Guided Holistic interactive (Holi) modules.
Specifically, RoCo aggregates the visual features into the row- and column-wise features corresponding two directional axes respectively. 
It offers a fine-grained matching behavior that perceives the associations between the linguistic features and two decoupled visual features to perform position reasoning over a hierarchical space.
Holi integrates features of the two modalities by a cross-modal attention mechanism, which suppresses the irrelevant redundancy under the guide of positioning information from RoCo.
Thus, with the incorporation of RoCo and Holi modules, CPRN captures the visual details of position reasoning so that the model can achieve more accurate segmentation.
To our knowledge, this is the first work that explicitly focuses on position reasoning modeling.
We also validate the proposed method on three evaluation datasets. It consistently outperforms existing state-of-the-art methods. 
\end{abstract}

%Besides, a Guided Holistic interactive (Holi) module is proposed for the global information. The module integrates the Roco's feature maps to ensure the integrity of representations.
%Besides, to ensure the integrity of the global information, a Guided Holistic interactive (Holi) module is proposed to preserve the holistic visual feature map while fusing Roco's positioning information for more accurate segmentation. 
%, which divides the holistic feature map into row- and column-wise maps and integrates them separately with textual features.


%==========Introduction==========
\section{Introduction}
\begin{figure}
	    \begin{center}
	    \centering
		\includegraphics[width=1\linewidth]{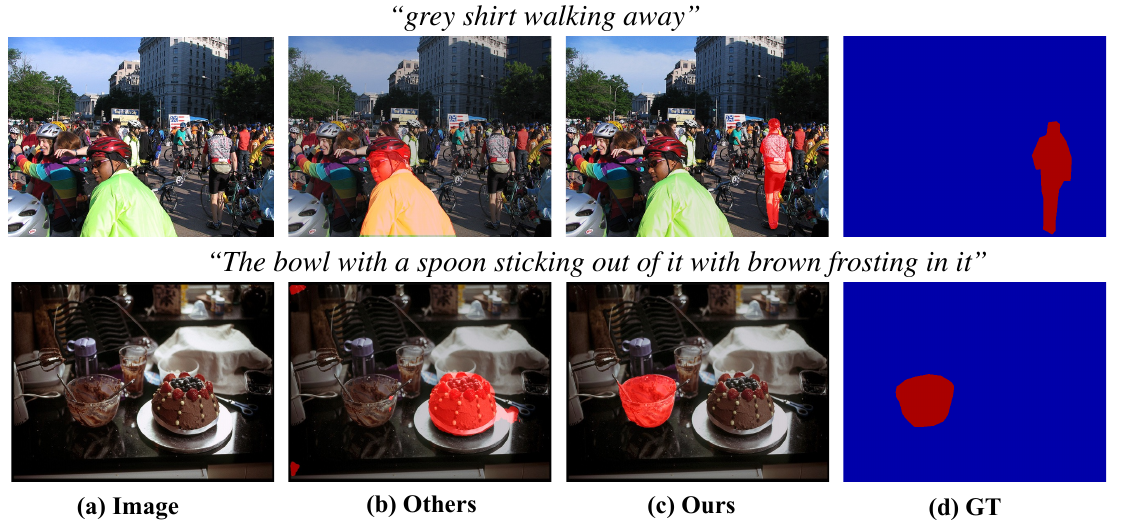}\\
		\end{center}
		\vspace{-0.3cm}
		\caption{\textbf{Comparison of visualization results between state-of-the-art methods and our proposed approach.} The first row shows that the existing methods are prone to positioning errors on some Non-Salient targets with the small-scale. The second row shows that for some complex referring expressions, the reasoning ability of the previous models is not enough to position the target accurately.
		}\label{fig:1} 
		%\vspace{-0.5cm}
\end{figure}

\IEEEPARstart{R}{eferring} image segmentation (RIS) aims to predict a pixel-level segmentation mask in the image corresponding to entity referred by the natural language expression. As shown in Fig.~\ref{fig:1}, it can identify the entities of interest by the description of free-form referring expressions, which are not restricted to pre-defined object categories. RIS requires the algorithms to explore the relationship between language and vision so that the style of referred entity can be more flexible than traditional segmentation tasks. Hence, RIS can be regarded as an open-ended task and has a wide range of potential applications in interactive image editing and human-robot interaction, etc. It has attracted the attention of many researchers in the intersection of vision and language.

\begin{figure*}
	\begin{center}
	\centering
		\includegraphics[width=1\linewidth]{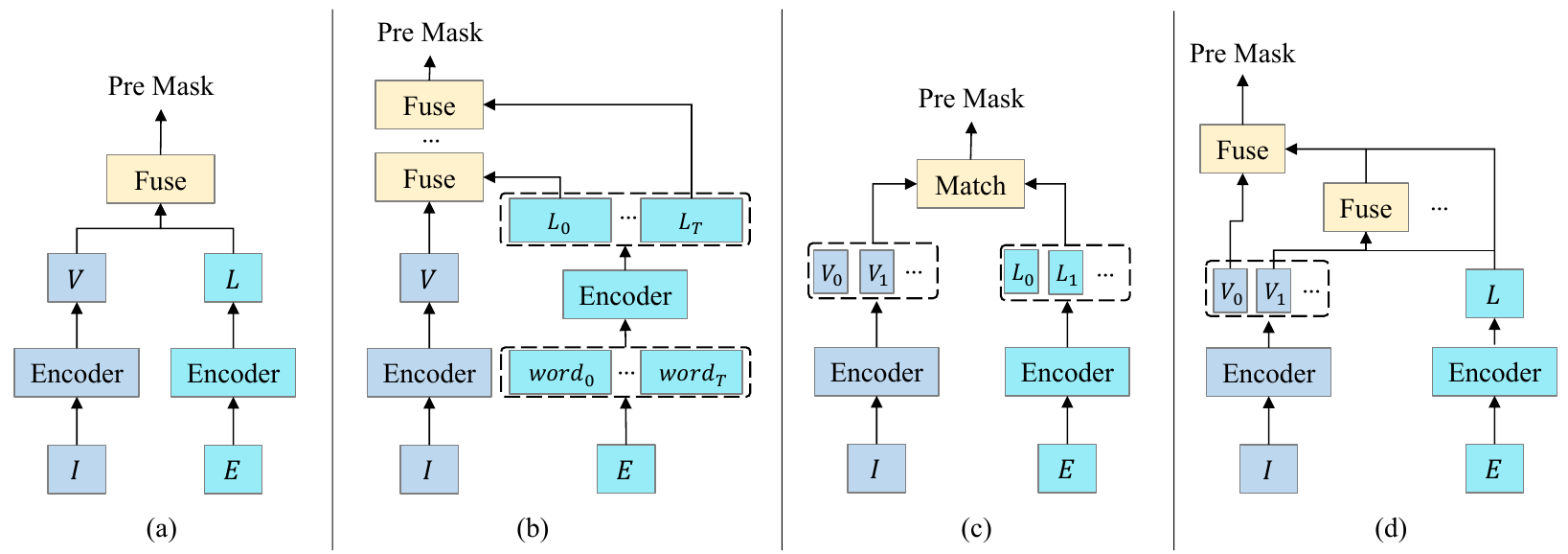}\\
	\end{center}
	\vspace{-0.4cm}
	\caption{ \textbf{Conceptual Comparisons of typical architectures.} (a) Simply combining the two modals of information. (b) Dividing the referring expression and fusing it with visual features progressively. (c) Exploring the relationship between visual and textual features by matching their attributes. (d) Adopting inference module in a multi-semantic-level progressively-fusion network architecture.
	%, which type our method belongs to. 
	$I$, $E$, $V$ and $L$ denotes image, natural language expression, visual feature and texture feature, respectively.
	}\label{fig:2} 
 \vspace{-0.2cm}
\end{figure*}

Since RIS involves visual and linguistic domains, it is challenging especially in modeling the fine-grained interactions and aligning implicit relationships among the two modalities. As shown in Fig.~\ref{fig:2}, existing works can be roughly divided into four types according to the network structure.
(a) A straightforward way to extract contextual knowledge to produce the final result via a simple concatenation-convolution scheme such as dynamic filters, LSTM and attention mechanism. This solution~\cite{2016Segmentation,2018RRN} aggregates the visual and linguistic features without a deep understanding, which could not effectively explore the relationships between the two modalities.
% without deep data mining and cyclic progressive fusion. For example, \cite{2016Segmentation} firstly proposes to model the image and the language independently and fuses them by concatenation. RRN~\cite{2018RRN} focuses on refining segmentation via multi-scale feature fusion and incorporates multi-scale information gradually by convolutional LSTM.
(b) Another line of works~\cite{2017mLSTM, 2018Dynamic} process each word in the referring expression to learn cross-modal interaction in a sequential manner. However, they consider each word as an equal contribution. This may have trouble distinguishing the target with long referring expressions.
(c) Alternative works~\cite{MAttNet,2019BMVC} establish several attributes (object, location and relationship to other objects) of referring expressions to improve the scores of cross-modal matching. This design help refine the segmentation results but lacks global context information and relies on the proposals generated by object detectors.
% MAttNet~\cite{MAttNet} decomposes the expression into three phrase embeddings (subject appearance, location, and relationship to other objects), uses them to trigger visual modules to compute matching scores. \cite{2019BMVC} finds consistency between the query and the generated caption from the image.
(d) At last, a series of works have been proposed to progressively integrate contextual information at multiple levels. LAVT~\cite{yang2022lavt} fuses the linguistic and visual features into each stage of the network and captures segmentation masks with a lightweight decoder. BRINet~\cite{2020Bi} considers the interaction through a bi-directional cross-modal attention module that uses both visual and textual guidances to capture their dependencies, realizing the compatibility between vision and textural features. They focuses on utilizing the inference module to enhance the visual and textual interaction achieves the best results. Nevertheless, all these methods do not directly focus on the issue of position reasoning and fail to locates the referred entity from background.

% Visual reasoning with language information is essential to segment the object region accurately for RIS.

% However, RIS is inherently a challenging task, which not only requires to explicitly locate the target entities in the image, but also enables efficient cross-modality feature alignment and interaction for fine-grained segmentation.
% Given an image and a natural language expression, 
Most previous works tackle the referring problem utilizing efficient cross-modality feature interaction to explore semantic contextual representations. Specifically, mainstream frameworks firstly extract visual and linguistic features respectively, and then introduce diverse operations to solve the interactive learning. Although these methods have achieved remarkable performance, the limitation of them is that the global context modelings still lack sufficient fine-grained visual concepts which is essential for position reasoning.
Fig.~\ref{fig:1} shows the visualization examples in which the segmentation masks remain unsatisfactory because of the distraction of background and non-salient objects.
From the perspective of human cognition, the RIS model focuses on positioning the entity regions that well match the expressions, and then refine the precise segmentation. The fine-grained semantic features help the model distinguish the referred entity from other analogs.

%Fig.~\ref{fig:2} shows a comparison of multi-modal fusion architectures.

%Recently, some works have noticed the referent positioning issue and tried to solve it via the transformer architecture. 
%Recently, some works have also tried to introduce transformer architecture to significantly improve the performance of RIS model.
Recently, the Transformer has achieved great success in the area of Natural Language Processing and Computer Vision.
The state-of-the-art RIS methods introduce Transformer architectures to strengthen the ability of multi-modal feature fusion and global information modeling. VLT~\cite{ding2021vision} uses a transformer to build a network with an encoder-decoder attention mechanism to enhance global contextual information. LAVT~\cite{yang2022lavt} utilizes the multi-stage design in the Swin Transformer to form a hierarchical language-aware visual coding scheme. CRIS~\cite{wang2022cris} leverages the pre-trained model CLIP~\cite{radford2021learning} and contrastive learning strategy to achieve text-pixel alignment. Although the transformer can bring a certain performance improvement to the RIS model, the challenge of position reasoning still exists and has not been well solved.

\begin{figure}
	\begin{center}
	\centering
    \includegraphics[width=1\linewidth]{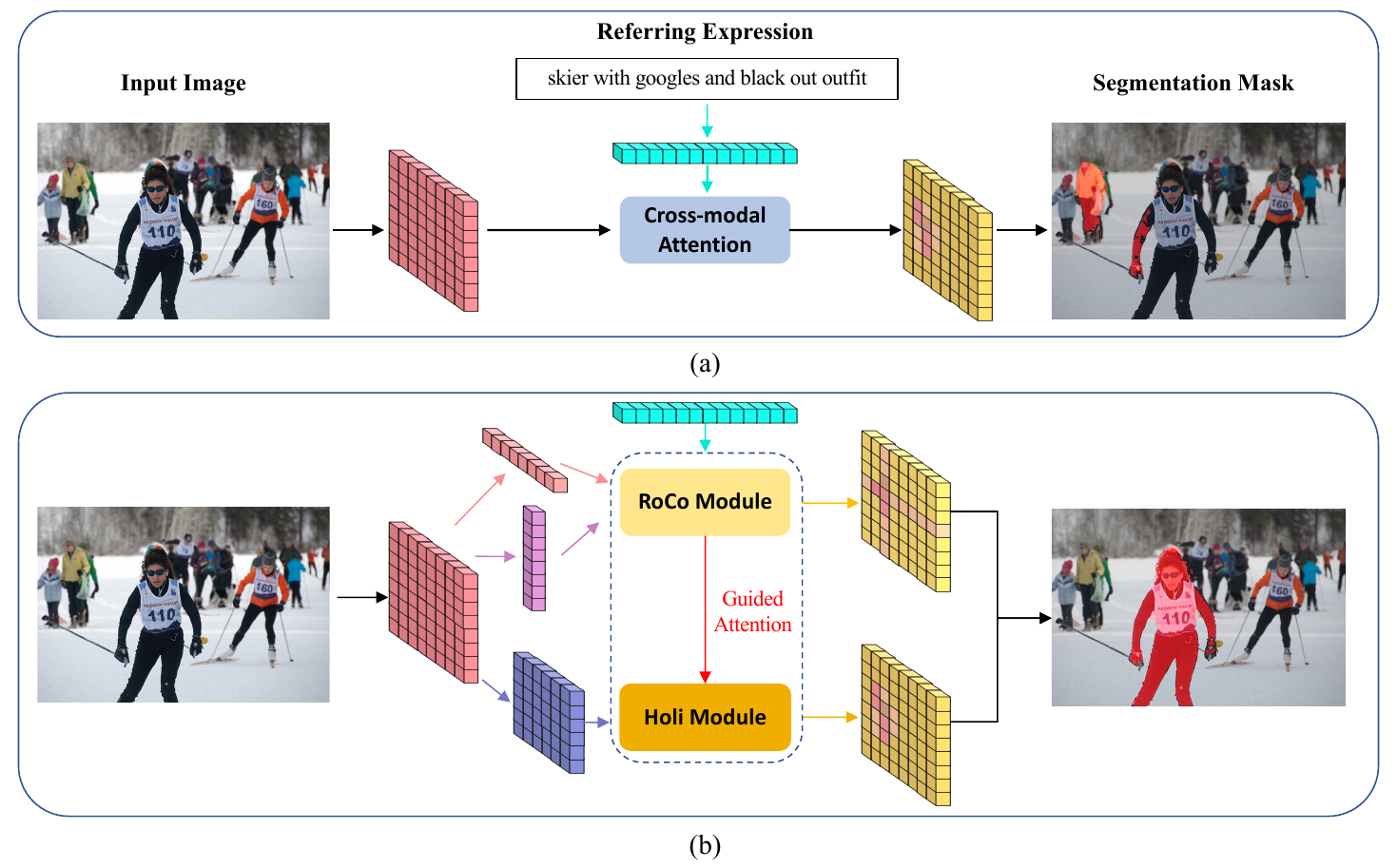}\\
		\end{center}
		\vspace{-0.4cm}
		\caption{\textbf{Illustration of the difference between (a) previous methods and (b) our model.} Previous works apply the holistic feature map in Cross-modal Attention. Differently, we use two parallel branches. The above one divides the holistic feature map into horizontal and vertical maps, and the lower one keeps the holistic feature map. They are fused with the textual feature, respectively, getting a better location for the referent.
		}\label{fig:3}
  \vspace{-0.2cm}
\end{figure}

In this paper, we address the problem of position reasoning in RIS and propose a Collaborative Positioning Reasoning Network (CPRN) for leveraging the hierarchical context of images for position reasoning. 
As illustrated in Fig.~\ref{fig:3}, in our model, the features passed through two parallel pathways can capture the local and global information for accurate localization and fine-grained segmentation. In  detail, the Row-and-Column interactive (RoCo) module generates the correlation between horizontal and vertical feature maps with linguistic features. The Guided Holistic interactive (Holi) module keeps the holistic feature map to ensure the integrity of global information. Meanwhile, a global guidance path directs the RoCo's positioning information into the Holi module to enhance entity perception reasoning and suppress the irrelevant redundancy from background. The output features of RoCo and  Holi are merged via a Feed Forward Network (FFN). Finally, we devise a Multi-Scale decoder to aggregate multi-level features  for accurate referring segmentation.
    
In summary, this paper makes the following contributions:
\begin{itemize}
\item We propose a novel Collaborative Positioning Reasoning Network (CPRN) to explicitly settle the position reasoning issue in RIS. And the proposed CPRN can be used as a flexible block adaptable to any inference-based framework.

\item We propose a Row-and-Column interactive (RoCo) module to explicitly locate the referent by dividing the holistic feature map into row- and column-wise maps and integrating them separately with textual features.

\item We propose a Guided Holistic interactive (Holi) module to retain a comprehensive perception of all pixels in an image, for fine-grained segmentation. Furthermore, the global guidance path is designed to enhance the localization of Holi by incorporating the RoCo's positioning information.
    
\item Extensive experiments on all three challenging datasets show that the proposed CPRN plays an important role in improving the positioning performance of referring image segmentation. And our model achieves superior performance compared to state-of-the-art methods.
\end{itemize} 

%\item We propose to decompose the overall visual features into vertical and horizontal directions for the task of reference image segmentation, for the first time. In the multi-modal feature fusion process, we process feature maps separately along two dimensions, applying row and column visual features to interact with textual features, respectively.
%\item Our proposed localization module is a flexible module that can be embedded on any model framework.

%========= related work =============
\section{Related Work}

\subsection{Referring Image Segmentation.}
Given an image and a natural language expression, the goal of Referring image segmentation is to produce a segmentation mask in the image corresponding to entities referred by the natural language expression. The RIS task is firstly introduced in~\cite{2016Segmentation}, which directly concatenates both visual and textual features to generate the final mask. RRN~\cite{2018RRN} considers the multi-scale semantics in the visual encoding step and employs ConvLSTM~\cite{2015Convolutional} in the feature fusion step. Later, word attention~\cite{2018Key} extracts keywords in the image regions to suppress noises in the referring expression and highlight the target object. 
RMI~\cite{2017mLSTM} directly combines visual features with each word feature from a language LSTM to recurrently refine segmentation results. 
DMNet~\cite{2018Dynamic} utilizes a dynamic filter for each word to further enhances this interaction. Further, relation inference is applied to capture visual and textual modalities. 
    
With the application of the attention mechanism more and more widely, some work uses the attention mechanism to extract visual content corresponding to language expression.
STEP~\cite{2019STEP} emphasizes the attention from image to word by computing dependencies between each visual region and each word, to guide the segmentation recurrently. 
CMSA~\cite{2019Cross} is exploited in respectively to capture global interaction information between image regions and words via Cross-modal self-attention. 
CMPC~\cite{huang2020CMPC} firstly employs entity and attribute words to perceive all the related entities. Then, the relational words are adopted to highlight the correct entity, as well as suppress other irrelevant ones by multi-modal graph reasoning. 
BRINet~\cite{2020Bi} uses both visual and linguistic guidances to capture the dependencies between multi-modal features. LSCM~\cite{2020Linguistic} models interaction between visual and textural information under the guidance of DPT-WG \cite{2014DPT}.
ReSTR~\cite{kim2022restr} is the first convolution-free architecture for RIS, unifying two different modal network topologies with Transformer.
CRIS~\cite{wang2022cris} uses  the pre-trained model CLIP~\cite{radford2021learning} and contrast learning strategies to achieve text pixel alignment. MaIL~\cite{li2021mail} introduces a new modal information mask mode and designs a simpler encoder-decoder pipeline and a mask-image-language three-mode encoder.
LAVT~\cite{yang2022lavt} fuse the linguistic and visual features into each stage of the network and captures segmentation masks with a lightweight decoder.
    
However, these works merely adopt holistic visual information in multi-modal interaction, leading to inaccurate object location. In this work, we introduce a collaborative position reasoning method by row-and-column interaction, in addition to holistic multi-interactive inference, and achieve satisfied segmentation results. 
    
\subsection{Multi-modal Interaction.}

A lot of multi-modal interaction researchers are interested in combining natural language processing with visual understanding. 
At first, ~\cite{2009Ngiam} demonstrates that if relevant data from different modalities is available at training time, better features can be learned. 
TFN~\cite{TFN}, LMF~\cite{LMF}, and T2FN~\cite{T2FN} are proposed to capture both intra- and inter-modal dynamics simultaneously. 
MulT~\cite{2019Multimodal} aligns data from different modalities implicitly, which leverages cross-modal attention modules for each modality on a high level, and each of them is responsible for aligning the target modality vector with the complementary modal vector.
\cite{2019Adaptive} introduces Auto-Fusion and GAN-Fusion learning to compress information from different modalities while preserving the context and GAN-Fusion regularizes the learned latent space given context from complementing modalities, making the network decide the fusion manner. 
MCF~\cite{MCF} puts forward reshaping feature vectors into circulant matrices and defining two types of interaction operations between vectors and matrices. 
\cite{ChannelAndPixel} realizes bidirectional multi-layer fusion from both channel-level and pixel-level through two fusion operations, which can strengthen the multi-modal feature interactions across channels as well as enhance the spatial feature discrimination. For a given query (image or language), ~\cite{chen2020uniter} simply considers the keys and values from all input tokens, it just merges the input from both ways, this multi-modal attention is called Merge attention.  
~\cite{lu2019vilbert} approach is that given a query from one modality (e.g., image), keys and values can only be obtained from another modality (e.g., language), this multi-modal attention is called co-attention. These methods for multi-modal interaction are based on holistic feature maps.
    
In this paper, we designed a Row-and-Column interactive (RoCo) module, which decompose the holistic feature map and used row- and column-wise information to interact with textual feature, respectively, to establish the local association between visual and linguistic patterns.
    
\begin{figure*}
\begin{center}
\centering
	\includegraphics[width=1\linewidth]{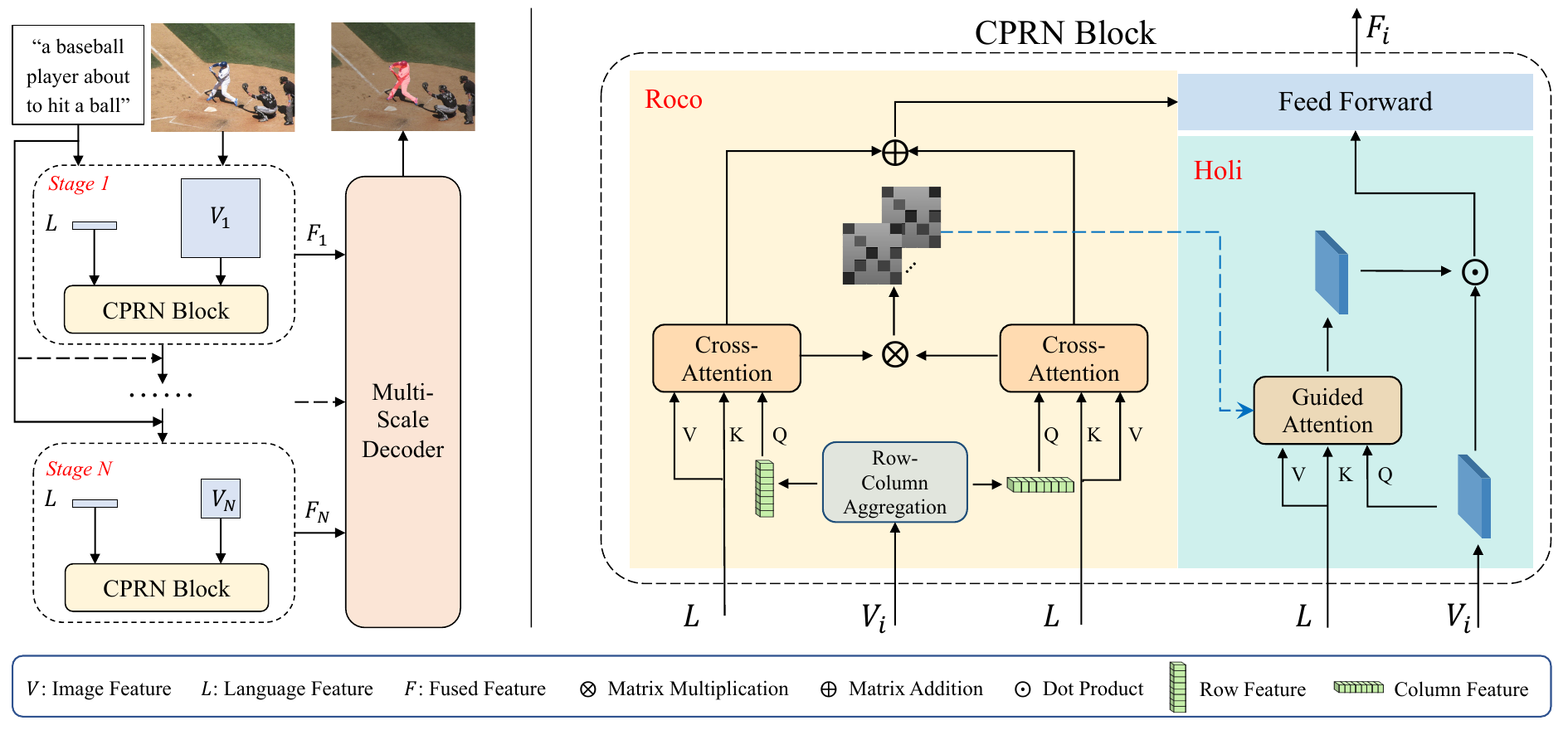}\\
\end{center}
\vspace{-0.5cm}
\caption{\textbf{The overall architecture of our method.} The CPRN block directly fuses features from two modal inputs of image $\bf{V}_{i}$ and text $\bf{L}$ to generate cross-modal feature representations. It includes RoCo interactive module and Holi interactive module. The multi-scale decoder generates segmentation results based on the interactive representation of multi-modal features at different stages.}
\label{fig:4} 
\end{figure*}
    
\subsection{Location Mechanism.}
    %CCNet
    %In the task of semantic segmentation, the problem of localization cannot be ignored. At present, a little work has been done on localizing target reference objects. ~\cite{liu2021few} uses the prior extractor to extract the prior, and then uses the prior to generate a prior region map for the query image, which is used to locate the object. LTS~\cite{jing2021locate} locates the objects referred to by linguistic expressions through simple correlation filtering and unified attention based block transformers.

    %Similar to our work is CCNet~\cite{huang2019ccnet}. CCNet obtains dense contextual information for semantic segmentation through two recurrent cross-attention modules (RCCA modules) and reduces the amount of computation. Different from CCNet, we propose two attention mechanisms, global and local, in parallel, where the local attention mechanism assists the global attention mechanism through two pathways.
In semantic segmentation tasks, in addition to multimodal feature fusion, the problem of locating reference images cannot be ignored. At present, some work has been done in locating target reference objects. ~\cite{liu2021few} uses the prior extractor to extract the prior, and then uses the before to generate a prior region map of the query image, which is used to locate objects. MCN~\cite{2020Multi} jointly learns two tasks, Citation Representation Comprehension (REC) and Segmentation (RES). To address the problem of conflicting predictions between the two tasks, he proposed an adaptive soft non-localization suppression (ASNLS) design, a post-processing method that suppresses responses in irrelevant regions in the RES based on the predictions of RECs. LTS~\cite{jing2021locate} proposes a localization module to obtain the corresponding visual content of the expression and uses the obtained object prior as the visual localization guidance for the subsequent segmentation module. Unlike the localization module, which includes two forms of simple filters and Transformers, the proposed CPRN block locates the visual area that responds to the linguistic expression by the row-column position information.
There is a class of methods to obtain object localization information by using additional external sources, such as MAttNet~\cite{2018MAttNet} and lang2seg~\cite{2019Referring}. These two methods use Mask R-CNN~\cite{2017Mask} to pre-process and post-process the image when segmenting the image. Although Mask R-CNN provides localization and segmentation of objects in images, greatly improving the performance of the model, our collaborative positioning and reasoning network performs much better on the three benchmark datasets, demonstrating the superiority of our approach in positioning. In addition, the idea of CCNet~\cite{huang2019ccnet} is somewhat similar to ours and it obtains dense contextual information for semantic segmentation through two recurrent cross-attention (RCCA) modules, which aggregates associated information via rows and columns. Different from CCNet, we design two parallel interactive modules, Roco and Holi, where Roco leverages the row and column information to explicitly locate the referent, and Holi utilizes the global image information for fine segmentation.

%===========Methodology==========
\section{Methodology}
Fig.~\ref{fig:4} illustrates the overall architecture of our Collaborative Positioning Reasoning Network, which integrates the proposed CPRN block for referring image segmentation. We first elaborate on the motivation of our approach in Sec.~\ref{sec:Motivation}.
Given an image and a natural language expression as input, we extract the visual and linguistic features on different semantic levels, respectively (Sec.~\ref{sec:Feature Extraction}). 
Then, they are fused and fed into the CPRN inference block (Sec.~\ref{sec:Collaborative Positioning and Joint Reasoning scheme}), which is composed of two modules, to highlight the referent entities. 
One is the Row-and-Column interactive (RoCo) module, the other is the Guided Holistic interactive (Holi) module. 
After that, the two pathways are merged using a Feed Forward Network (FFN) to enhance the reasoning features. Finally, the Multi-Scale Decoder module (Sec.~\ref{sec:Multi-Scale Decoder module}) is used to perform the different stage feature fusion and refine the final segmentation mask.  

\subsection{Motivation}

\label{sec:Motivation}
It is essential for RIS task to mine relation information between vision and language via feature interaction. 
Some works~\cite{2019Cross,2018RRN,2020Linguistic} consider the multi-scale information to find the referent. 
Since they only consider holistic visual information, inappropriate segmentation results exist. A main problem is that they cannot accurately locate the object. 
We propose the Collaborative Positioning Reasoning Network (CPRN) utilizing two parallel pathways to sufficiently aggregate object position (RoCo module) while capturing holistic information (Holi module) between visual and textual modalities, as illustrated in Fig.~\ref{fig:4}.  
For solving the referent positioning issue, we decompose the visual feature into row- and column-wise features, which will interact with the textual features separately, to locate the object in both horizontal and vertical directions. 
The multi-modal features of the two directions will assign the location of the referent object. Meanwhile, the positioning effect of the RoCo module will also guide the Holi module, helping it to more accurately locate and segment the referent. By the mutual enhancement between the RoCo and Holi modules, our method can perform reliable joint reasoning, which greatly improves the localization and segmentation of referent entities.
    
\begin{figure*}
	\begin{center}
	\centering
		\includegraphics[width=0.9\linewidth]{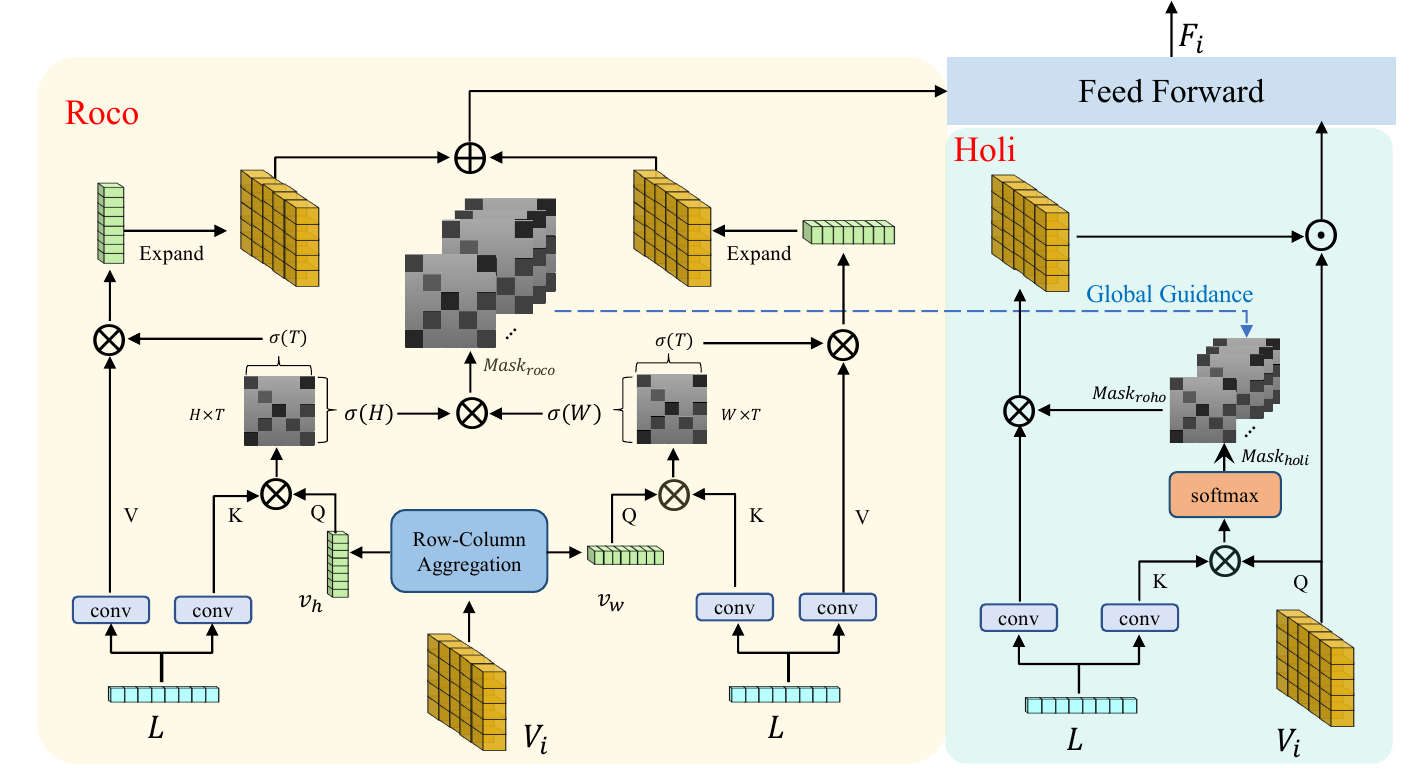}\\
	\end{center}
	\vspace{-0.3cm}
	\caption{\textbf{Illustration of our Collaborative Positioning Reasoning Network (CPRN).} First, the visual features $\bf{V}_i$ generate vertical features and horizontal features via the Row-Column Aggregation. These two visual features are obtained through cross attention layers and it generates the multi-modal features with semantic information $\bf{v}_{h}^{hw}$ and $\bf{v}_{w}^{hw}$. Then, the final output of RoCo module are obtained through expand and addition operations. At the same time, $\bf{mask}_{roco}$ with positioning information is calculated, which guides the Holi module through the global guidance pathway for fine-grained segmentation. The Holi module utilizes $\bf{Mask}_{roco}$ and $\bf{Mask}_{holi}$ to generate the final output via a designed Guided Attention layer. Expand represents bilinear interpolation operation, and for convenience of representation, we ignore some residual connections.}
	\label{fig:5} 
\end{figure*}

\subsection{Feature Extraction}

\label{sec:Feature Extraction}
Taking an image $I$ and a referring expression $E$ with $T$ words as input, we firstly use the Swin Transformer~\cite{liu2021swin} to extract visual features at different stages. 
Let $V_i \in \mathbb{R}^{ H_i \times W_i \times C_i }$, $i \in \{2, 3, 4, 5\}$, denotes the visual features, corresponding to the 1st, 2nd, 3rd and 4th stages of Swin Transformer network, where $H_i$, $W_i$ and $C_i$ are the dimensions of height, width, and visual feature channels, respectively. Besides, the spatial coordinate features are used to capture more spatial information. For each stage, we also define an 8-D spatial coordinate feature denoted as $P_i \in \mathbb{R}^{H_i \times W_i \times 8}$, $i \in \{2, 3, 4, 5\}$ at each spatial position as the implementation in~\cite{2017mLSTM}. 
Then, a new fused visual feature $\mathbf{V}_i$ is obtained by concatenating the visual feature $V_i$ and the spatial coordinate feature $P_i$ followed by a 1$\times$1 convolution layer. 
We denote a single level of fused visual features as $\mathbf{V}$ for ease. 
Next, the linguistic features $\mathbf{L} = \{\mathbf{L}_1, \mathbf{L}_2, ..., \mathbf{L}_T\}$, $\mathbf{L}_i \in \mathbb{R}^{d_l}, i \in \{1, 2, ..., T\}$ is extracted with a language encoder BERT~\cite{devlin2018bert},
where $d_l$ and $T$ denote the number of channels and the number of words. 
After that, the visual features $\mathbf{V}$ and linguistic features $\mathbf{L}$ are fed into our proposed Collaborative Positioning Reasoning Network (CPRN) which depicted in Sec.~\ref{sec:Collaborative Positioning and Joint Reasoning scheme}.

\subsection{Collaborative Positioning Reasoning Network}
\label{sec:Collaborative Positioning and Joint Reasoning scheme}
Rather than fusing the visual and linguistic features directly in previous works, the proposed Collaborative Positioning Reasoning Network (CPRN) pays more attention to the positioning of referent entities and designed the Row-and-Column interactive (RoCo) module, which realizes the positioning target by perceiving the row- and column-wise local features of the image. As illustrate in Fig.~\ref{fig:5}, it also designs the Guided Holistic interactive (Holi) module, which realizes the accurate segmentation of referents by perceiving the global features of the image. Furthermore, the Feed Forward Network (FFN) is designed to merge the two parallel pathways, enabling joint reasoning, making the features used for final segmentation more reliable.

\subsubsection{Row-and-Column interactive module}
\label{sec:Row-and-Column interactive module}
In the positioning path, the network firstly decomposes the visual feature map $\mathbf{V_i}$ into two parts, \emph{i.e.} a row-wise feature and a column-wise feature, corresponding to the horizontal and the vertical directions, respectively.
After that, it leverages to interact among the two visual features and the linguistic features, as shown in Fig.~\ref{fig:5}. For the convenience of representation, we remove the $i$ subscript of all variables.
\par Specifically, the row-wise and column-wise visual features are obtained by the Row-Colunm Aggregation operations which execute average pooling on $\mathbf{V}$, with pooling kernel of size $1\times W$ and $H\times 1$. And each of these two features after pooling is implemented with 1$\times$1 convolution layer and followed by the GeLU function adding nonlinearity:
\begin{equation}
	\begin{split}
		\mathbf{v}_h &= GeLU\left( \mathbf{w}^1_h\left ( Avg\_pool_h \left ( \mathbf{V} \right ) \right )+\mathbf{b}^1_{h})\right)  , \\
		\mathbf{v}_w &=  GeLU\left(\mathbf{w}^1_w\left ( Avg\_pool_w \left ( \mathbf{V} \right ) \right )+\mathbf{b}^1_{w}\right),
	\end{split}
\end{equation}\label{eq:1}
where $\mathbf{v}_h \in \mathbb{R}^{H \times C_{h}}$ denotes the row-wise feature, and $\mathbf{v}_w \in \mathbb{R}^{W \times C_{w}}$ denotes the column-wise feature, $C_{h} = C_{w} = C$. $H$ is the height of vertical feature, $W$ represents the width of horizontal feature, $C_{h}$ and $C_{w}$ define the number of channels. 
From the linguistic feature $\mathbf{L}$, it generates word vectors, $\mathbf{word}_{h_k} \in \mathbb{R}^{T \times d_{h}}$, $\mathbf{word}_{h_v} \in \mathbb{R}^{T \times d_{h}}$, $\mathbf{word}_{w_k} \in \mathbb{R}^{T \times d_{w}}$, $\mathbf{word}_{w_v} \in \mathbb{R}^{T \times d_{w}}$, through four 1$\times$1 convolution layers:
\begin{equation}
	\begin{split}
		\mathbf{word}_{h_k} &=  \mathbf{w}^2_{h_k}\left ( \mathbf{L} \right )+\mathbf{b}^2_{h_k} , \\
		\mathbf{word}_{h_v} &=  \mathbf{w}^2_{h_v}\left ( \mathbf{L} \right )+\mathbf{b}^2_{h_v} , \\
		\mathbf{word}_{w_k} &=  \mathbf{w}^2_{w_k}\left ( \mathbf{L} \right )+\mathbf{b}^2_{w_k} , \\
		\mathbf{word}_{w_v} &=  \mathbf{w}^2_{w_v}\left ( \mathbf{L} \right )+\mathbf{b}^2_{w_v} . 
	\end{split}
\end{equation}\label{eq:2}

Then, those word vectors are passed to the row-wise inference and the column-wise inference branches, respectively, to fully capture the two directional interactions. In detail, it feeds the row-wise and column-wise visual features and the corresponding word vectors into two cross-attention mechanisms separately, to calculate the language perception of the row-and-column level pixels in the image.
Since the implementations of these two cross-attention mechanisms are the same, for simplicity, we only take the row-wise inference as an example. Specifically, the common cross-attention mechanisms are utilized to learn the row-wise influence by feeding the vertical visual feature $\mathbf{v}_h$ to query the linguistic feature $\mathbf{word}_{h_k}$ and generate the vertical linguistic feature. $Attention$ is the simple Scaled Dot-Product Attention mechanism and can be expressed by:
\begin{equation}
	\begin{split}
		Attention(Q, K, V) &= softmax \left ( \frac{ Q K^\top}{\sqrt{d_k}} \right ) V.
	\end{split}
\end{equation}\label{eq:3}

After obtaining the vertical linguistic feature which have the same shape as $\mathbf{v}_{h}$, we combine them to produce a set of vertical multi-modal feature maps $\mathbf{v}_{h}^{Att}$ via element-wise multiplication. Formally,
\begin{equation}
	\begin{split}
	    \mathbf{v}_{h}^{Att} = Attention\left (\mathbf{v}_{h}, \mathbf{word}_{h_k}, \mathbf{word}_{h_v} \right ) \odot \mathbf{v}_{h},
	\end{split}
\end{equation}\label{eq:4}
where $\mathbf{v}_{h}^{Att} \in \mathbb{R}^{H \times C_{h}}$, $\odot$ denotes element-wise multiplication. In the same way, we use another cross-attention layer to generate the horizontal multi-modal feature maps $\mathbf{v}_{w}^{Att}\in \mathbb{R}^{W \times C_{w}}$. Finally, we use the Bilinear Interpolation to resize $\mathbf{v}_h$, $\mathbf{v}_w$, $\mathbf{v}_h^{Att}$, $\mathbf{v}_w^{Att}$ to the scale of the original image, which are added up as the output of the Row-and-Column interactive module for further fusion:
\begin{equation}
	\begin{split}
		\mathbf{v}_{hw}^{all} = B \left( \mathbf{v}_{h} \right) + B \left( \mathbf{v}_{w} \right) + B \left( \mathbf{v}_{h}^{Att} \right) + B \left( \mathbf{v}_{w}^{Att}\right), \\
	\end{split}\label{eq:5}
\end{equation}
where $\mathbf{v}_{hw}^{all} \in \mathbb{R}^{H \times W \times C}$ and $B$ denotes the Bilinear Interpolation layer. It is worth noting that in order to preserve the row-wise and column-wise visual features of the image, we also use Bilinear Interpolation for $\mathbf{v}_h$, $\mathbf{v}_w$ and add them to the final output, which is not shown in the above figures.

Furthermore, to enable the positioning effect of the RoCo module to guide the Holi module, we also design a global guidance path that utilizes the learned horizontal and vertical attention maps to build the global perception of the image, giving the Holi module a referent location prior $\mathbf{Mask}_{roco} \in \mathbb{R}^{H \times W \times T} $, which can be formulate as
\begin{equation}
	\begin{split}
	    \mathbf{Mask}_{roco} =
	    \frac{\mathbf{e}_{h} * \mathbf{e}_{w}^\top}{{\sum_{HW} \left ( \mathbf{e}_{h} * \mathbf{e}_{w}^\top \right )}} , \\
	    %\frac{\sum_{i=0}^\top \left ( \mathbf{e}_{h} * \mathbf{e}_{w}^\top \right )}{{T\sum_{HW} \left ( \mathbf{e}_{h} * \mathbf{e}_{w}^\top \right )}} , \\
	    \mathbf{e}_{h} = \sigma \left ( \frac{ \mathbf{word}_{h_k}  \mathbf{v}_h^\top}{\sqrt{d_h}} \right ) , \\
	    \mathbf{e}_{w} = \sigma \left ( \frac{ \mathbf{word}_{w_k}  \mathbf{v}_w^\top}{\sqrt{d_w}} \right ), \\
	\end{split}\label{eq:6}
\end{equation}
where $\mathbf{e}_h \in \mathbb{R}^{H \times T}$, $\mathbf{e}_w \in \mathbb{R}^{W \times T}$ represent the vertical and horizontal attention maps, respectively. $T$ denote the number of words and $\sigma$ denote the $softmax$ function. The generated $\mathbf{Mask}_{roco}$ will be used to guide the Holi module which is depicted in Sec.~\ref{sec:Guided Holistic interactive module}. 
In the above process, some small regions in the holistic feature map can be enhanced. In other words, some small-scale non-salient objects in the image could be explicitly located like salient objects, resulting in more accurate segmentation masks.

\subsubsection{Guided Holistic interactive module} 

\label{sec:Guided Holistic interactive module} 
As illustrated in Fig.~\ref{fig:5}, our Guided Holistic interactive (Holi) module establishes the attention correlations between the holistic visual features and the linguistic features. Like previous methods, we maintain the scale of the visual feature map during the multi-modal feature interaction, which in turn captures the perception between the language words and the image pixels. In detail, an 1$\times$1 convolution layer are implemented to obtain the holistic visual feature $\mathbf{v}_g \in \mathbb{R}^{H \times W \times C}$, which can be formulate as
\begin{equation}
	\begin{split}
	    \mathbf{v}_g &= \mathbf{w}^3_g (\mathbf{V}) + \mathbf{b}^3_g. \\
	\end{split}
\end{equation}

In addition, from the linguistic feature $\mathbf{L}$, we also utilize two 1$\times$1 convolution layers to generate $\mathbf{word}_{g_k} \in \mathbb{R}^{T \times d_g}$ and $\mathbf{word}_{g_v} \in \mathbb{R}^{T \times d_g}$.
\begin{equation}
	\begin{split}
		\mathbf{word}_{g_k} &= \mathbf{w}^3_{g_k}\left ( \mathbf{L} \right )+\mathbf{b}^3_{g_k} , \\
		\mathbf{word}_{g_v} &= \mathbf{w}^3_{g_v}\left ( \mathbf{L} \right )+\mathbf{b}^3_{g_v} . \\
	\end{split}
\end{equation}

Then, a novel Guided Attention layer is designed to capture the multi-modal interactions between the linguistic features and the holistic visual features. Specifically, under the global guidance of the RoCo module, we first use a simple attention layer to calculate the holistic attention map $\mathbf{Mask}_{holi} \in \mathbb{R}^{H \times W \times T}$, and then fuse it with $\mathbf{Mask}_{roco}$ to generate the guided holistic attention map $\mathbf{Mask}_{roho} \in \mathbb{R}^{H \times W \times T}$, which could be defined as
\begin{equation}
	\begin{split}
    \mathbf{Mask}_{roho} = \frac{ \left ( \mathbf{Mask}_{roco} +  \mathbf{Mask}_{holi} \right ) }{2} , \\
	\end{split}
\end{equation}
\begin{equation}
	\begin{split}
	    \mathbf{Mask}_{holi} = \sigma \left ( \frac{ \mathbf{v}_g \mathbf{word}_{g_k}^\top}{\sqrt{d_g}} \right ) . \\
	\end{split}
\end{equation}

Finally, based on the guided holistic attention map $\mathbf{Mask}_{roho}$, we can get the holistic linguistic features, which have the same shape as $\mathbf{v}_g$, and then combine them to produce a set of holistic multi-modal feature maps $\mathbf{v}_{g}^{all} \in \mathbb{R}^{H \times W \times C}$ as the output of the Guided Holistic interactive module. Formally,
\begin{equation}
	\begin{split}
	    \mathbf{v}_{g}^{all} =  \left ( \mathbf{Mask}_{roho} * \mathbf{word}_{g_v}  \right ) \odot \mathbf{v}_{g},\\
	\end{split}
\end{equation}
where $*$ denotes matrix multiplication and $\odot$ denotes element-wise multiplication.

\subsubsection{Merging two pathways}
\label{sec:Merging two pathways}
In the following steps, a Feed Forward Network is utilized to fuse the outputs of these two parallel branches. Firstly, we use two convolution layers followed by ReLU nonlinearity to perform feature projection on $\mathbf{v}_{hw}^{all} $ and $\mathbf{v}_{g}^{all}$ and mathematically described as follows
\begin{equation}
	\begin{split}
		\mathbf{F}_{hw} &= ReLU\left ( \mathbf{w}_{hw}^{4}\left ( \mathbf{v}_{hw}^{all} \right )+\mathbf{b}_{hw}^{4} \right ),
	\end{split}
\end{equation}
\begin{equation}
	\begin{split}		
		\mathbf{F}_{g} &= ReLU\left ( \mathbf{w}_{g}^{4}\left ( \mathbf{v}_{g}^{all} \right )+\mathbf{b}_{g}^{4} \right ),
	\end{split}
\end{equation}
where $\mathbf{F}_{hw}, \mathbf{F}_{g} \in \mathbb{R}^{H \times W \times C}$. After that, the above two multi-model features are added up and follow by a feed forward network to generate the fused multi-modal features $\mathbf{F} \in \mathbb{R}^{H \times W \times C}$, involving the joint reasoning information learned by the RoCo and Holi modules.
\begin{equation}
	\begin{split}		
		\mathbf{F} &= update( FFN \left ( \mathbf{F}_{hw} + \mathbf{F}_{g} \right ) , \mathbf{V} ), \\
	\end{split}
\end{equation}
where $FFN$ denotes the traditional feed forword network which contains two linear projection layers and a ReLU nonlinear layer followed by dropout function. $update$ denotes the residual connection. In fact, $\mathbf{F}$ is the final output of our proposed CPRN. Under the cooperation and guidance between the RoCo module and the Holi module, our proposed CPRN block can better locate the referents and obtain more fine-grained results, making up for the previous models that only rely on the holistic visual feature map to locate the referents.

\subsection{Multi-Scale Decoder module}
\label{sec:Multi-Scale Decoder module}
As illustrate in Fig.~\ref{fig:4}, we combine the multi-modal features of different stages via a Multi-Scale Decoder module which progressively integrates these features $\{\mathbf{F}_{i}\}_{i=0}^{N}$ from high-level to low-level semantics as follows:
\begin{equation}
	\begin{split}		
		\mathbf{Y}_{i} &= Upsampling ( Proj \left ( [\mathbf{Y}_{i+1}, \mathbf{F}_{i+1}] \right ) ) . \\
	\end{split}
\end{equation}
where $Upsampling$ represents upsampling operation on the feature map via bilinear interpolation and $Proj$ indicates that the linear projection function is used to transform the channel dimension. Specifically, the output of final stage $\mathbf{Y}_{N}$ is equal to $\mathbf{F}_{N}$, and $N$ represents the number of different stages. At last, the final feature maps, $\mathbf{Y}_{1}$, are fed into an $1\times1$ convolution layer to produce a 2-D probability score map $y' \in \left( 0,1 \right)$ normalized with sigmoid function.
During training, a binary cross entropy loss function are utilized to calculate the loss between the predicted score map $y'$ and ground truth label $y$, which can be formulated as follow:
\begin{equation}
	\begin{split}		
	L = -\frac{1}{\mathcal{Z}} \sum^{\mathcal{Z}}_{n=0} [y\left( n \right) log(y'\left( n \right)) + (1-y\left( n \right))log(1-y'\left( n \right))]
	\end{split}
\end{equation}
where $n$ represents the n-th image pixel, and $\mathcal{Z}$ is the number of pixels in the input image. The detailed process of our CPRN block is expressed in algorithm~\ref{alg:Framwork}.

\begin{algorithm}[t]
    \caption{Framework of our CPRN.}
    \label{alg:Framwork}
    % \begin{algorithmic}[1]
    \KwIn{Images $I$, Language expression $E$;}
    \KwOut{Segmentation result $y'$;}
    
    Extracting the feature $\bf{V}$ with the help of Swin Transformer; 
    
    Extracting the feature $\bf {L}$ with the help of BERT;
    
    \For{stage $i \to N$}
    {
        $\mathbf{v}_h, \mathbf{v}_w = RoCo\_Aggregation\left(\bf{V}\right)$;
        
        $\mathbf{v}_{h}^{Att}, \mathbf{v}_{w}^{Att}= Cross\_Attention\left (\mathbf{v}_{h}, \mathbf{v}_{w};  \mathbf{L}\right)$;
        
        $\bm{v}_{hw}^{all} = B \left( \bm{v}_{h} \right) + B \left( \bm{v}_{w} \right) + B \left( \bm{v}_{h}^{Att} \right) + B \left( \bm{v}_{w}^{Att}\right)$;
        % Get the RoCo feature map $\bm{v}_{hw}^{all}$ by adding up the resized vertical feature maps $\bm{v}_{h}^{Att}$, oriental feature maps $\bm{v}_{w}^{Att}$, row-wise and column-wise visual feature;
        % $\bm{F}_{hw} = feature\_projection(\bm{v}_{hw}^{all})$
        
        Calculate the referent location prior $\mathbf{Mask}_{roco}$ through the global guidance path;
        
        % Calculate the RoCo feature map $\bm{F}_{hw}$ through RoCo attention and get the location attention map $\bm{Mask}_{roco}$
        Calculate the holistic attention map $\mathbf{Mask}_{holi}$ through a simple attention layer;
        
        Get the guided holistic attention map $\mathbf{Mask}_{roho}$ by fusing $\mathbf{Mask}_{roco}$ and $\mathbf{Mask}_{holi}$;
        
        % $\mathbf{v}_{g}^{all} =  \left ( \mathbf{Mask}_{roho} * \mathbf{word}_{g_v}  \right ) \odot \mathbf{v}_{g}$;
        $\mathbf{v}_{g}^{all} = Guided\_Attention\left (\mathbf{v}_{g}; \mathbf{L}| \mathbf{Mask}_{roho}\right)$;
        
        $\mathbf{F}_{hw}, \mathbf{F}_{g} = projection(\mathbf{v}_{hw}^{all}, \mathbf{v}_{g}^{all})$;
        
        $\mathbf{F} = update\left(FFN\left(\mathbf{F}_{hw} + \mathbf{F}_{g}\right), \mathbf{V}\right)$;
    }
    
    Get the fusion feature $\bf{F}$ through the multi-scale decoder;
    
    Calculate the Segmentation results $y'$ through a simple segmentation head;
    
    % \end{algorithmic}
\end{algorithm}

\begin{table*}[t]
\begin{center}
\caption{Comparison with state-of-the-art methods on three benchmark datasets using \emph{overall IoU} as metric. U: The UMD split. G: The Google split.}
\label{table:1}
\setlength{\tabcolsep}{2.5mm}{
\renewcommand{\arraystretch}{1.2}
% \resizebox{0.89\textwidth}{!}{
\begin{tabular}{l|ccc|ccc|ccc}
\hline
&  \multicolumn{3}{c|}{RefCOCO}                    
&  \multicolumn{3}{c|}{RefCOCO+}                                     
&  \multicolumn{3}{c}{Gref}  \\ 
\cline{2-10} 
\multirow{-2}{*}{} &  val & test A & test B & val & testA & testB &  val (U) &  test (U)  & val(G)  \\ 
\hline
\hline
RRN~\cite{2018RRN}   
&  {55.33} &  {57.26} & 53.93 &  {39.75}  &  {42.15} & 36.11  &  {-}  &  {-} & 36.45        \\
CSMA~\cite{2019Cross}
&  {58.32}  &  {60.61}    & 55.09 &  {43.76}  &  {47.60}    & 37.89  &  {-}  &  {-}        & 39.98 \\
BRINet~\cite{2020Bi}
&  {60.98}   &  {62.99}    & 59.21 &  {48.17}  &  {52.32}    & 42.11 &  {-}   &  {-}        & 48.04 \\
CMPC~\cite{huang2020CMPC}               
&  {61.36}   &  {64.53}  & 59.64 &  {49.56}   &  {53.44}  & 43.23 &  {-}       &  {-}      & 49.05 \\
LSCM~\cite{2020Linguistic}          
&  {61.47}  &  {64.99} & 59.55 &  {49.34}  &  {53.12} & 43.50  &  {-} &  {-}     & 48.05 \\
EFN~\cite{2021Encoder} 
&  {62.76}  &  {65.69} & 59.67 &  {51.50} &  {55.24} & 43.01 &  {-} &  {-}   & 51.93 \\
BUSNet~\cite{yang2021bottom}
&  {63.27} &  {66.41} & 61.39  &  {51.76} &  {56.87} & 44.13 &  {-} &  {-}    & 50.56 \\

VLT~\cite{ding2021vision}                
&  {65.65}   &  {68.29}  & 62.73 &  {55.50}   &  {59.20}  & 49.36 &  {52.99}   &  {56.65}  & 49.76 
\\
LTS~\cite{jing2021locate}                
&  {65.43}   &  {67.76}  & 63.08 &  {54.21}   &  {58.32}  & 48.02 &  {54.40}   &  {54.25}  & -     
\\
ReSTR~\cite{kim2022restr}              
&  {67.22}   &  {69.30}  & 64.45 &  {55.78}   &  {60.44}  & 48.27 &  {54.48}   &  {-}      & -  
\\
CRIS~\cite{wang2022cris}         
&  {70.47}   &  {73.18}  & 66.10 &  \underline{62.27}   &  {68.08}  & 53.68 &  {59.87}   &  {60.36}  & -  
\\
%MaIL     
%&  {70.13}   &  {71.71}  & 66.92 &  {62.23}   &  {65.92}  & 56.06
%&  {62.45}   & {62.87}  & 61.81 \\
LAVT~\cite{yang2022lavt} &  \underline{72.73}   & \underline{75.82} & \underline{68.79} &  62.14   & \underline{68.38} & \underline{55.10}   &  \underline{61.24}   &  \underline{62.09} & \underline{60.50}   \\ 
\hline
\hline
{CPRN (Ours)}      
&  {{\textbf{73.42}}}  &  {{\textbf{76.65}}}  & \textbf{70.84}  &  {\textbf{63.58}} 
&  {\textbf{69.44}}   & \textbf{55.84} &  \textbf{62.81}    &  {\textbf{64.25}}  & \textbf{60.92}               
\\ \hline
\end{tabular}}
\end{center}
\end{table*}

%==========Experiments==========
\section{Experiments}

\subsection{Datasets and Experiment Setup}

\begin{table*}
\begin{center}
\caption{Ablation studies on RefCOCO validation set. “\&” and “$\Vert$” represent the series and parallel connection of RoCo module and Holi module, respectively. "Holi*" represents a simple cross attention mechanism, and "Holi" represents our proposed guided holi interactive module.}\label{table:2}
\setlength{\tabcolsep}{2.5mm}{
\renewcommand{\arraystretch}{1.3}
\begin{tabular}{l| c| c| c| c| c| c| c}
\hline
\textbf{Method} & \textbf{P@0.5} & \textbf{P@0.6} & \textbf{P@0.7} & \textbf{P@0.8} & \textbf{P@0.9} & \textbf{Overall IoU} & \textbf{Mean IoU} \\  
\hline
\hline
{baseline (Holi*)} & 83.26  & 79.31  & 73.38  & 62.29  & 32.36  & 71.99 & 73.10   \\
RoCo & 79.27 &  74.68 &  67.35 & 50.45  & 15.11 & 66.63 & 67.61 \\
{RoCo \& Holi*} & 80.61 & 75.81  & 67.92 & 50.98 & 17.58 & 68.57 & 69.02 \\ 
{RoCo $\Vert$ Holi*} & 84.56 & 80.65 & 75.10  & 64.21 &  33.90 & 72.79 & 74.29 \\ 
{RoCo $\Vert$ Holi} &  84.58 & 81.21 & 75.91 & 64.28 & 34.04  & 72.96 & 74.48  \\
{   +FFN} & 84.66 & 81.49 & 76.21 &  65.23 & 35.00 & 73.12 &  74.60  \\
{   +ape (CPRN)} & 85.09 & 81.71 & 76.54 &  65.53 & 35.26 & 73.42 & 75.00   \\
\hline
%\hline
%{CPRJ (large)} & 85.61 & 82.22 & 73.13 & 66.85 & 36.28  &  73.97 &  75.46   \\ 
%\hline 
\end{tabular}}
\end{center}
\end{table*}

\textbf{Datasets:} 
We use three datasets to evaluate our method: RefCOCO~\cite{2016Modeling}, RefCOCO+~\cite{2016Modeling} and Gref~\cite{2016Generation}.

The RefCOCO~\cite{2016Modeling} set contains 19,994 images and 142,209 citation expressions for 50,000 objects obtained from the MS COCO dataset~\cite{2014Microsoft}, and the average length of Refcoco expressions is 3.61. The set annotation comes from the two-player game interaction~\cite{2014ReferItGame}, where two or more objects of the same object class appear in each image.

The RefCOCO+~\cite{2016Modeling} set contains 141,564 expressions for 49,856 objects in 19,992 images, and the average length of Refcoco+ is 3.53. These images were also collected from the MS COCO dataset, with a limitation that position words are not allowed in expressions. Refcoco and Refcoco+ do not limit the number of objects of the same category to 4, so containing some images with many objects of the same category, Refcoco, and Refcoco+ both average 3.9 same category objects per image. 

The Gref~\cite{2016Generation} set is also collected from the MS COCO dataset. There are 104,560 expressions involving 54,822 objects in 26,711 images. The expressions for this dataset are collected on Mechanical Turk through independent rounds, rather than using the two-player game. The expressions in Gref are longer and more complex than RefCOCO and RefCOCO+, with Gref containing an average of 8.43 words. We use the same split as in~\cite{2016Generation}, this dataset has two different splits, one is UMD and the other is Google, abbreviated as Gref-umd and Gref-google. Gref has an average of 1.63 same category objects per image. 

\textbf{Metrics:} Following previous works~\cite{2017mLSTM, 2019Cross, yang2022lavt, wang2022cris}, we adopt overall Intersection-over-Union (\emph{Overall\ IoU}) and \emph{Pre@X} as our evaluation metrics. Given the predicted segmentation mask and the ground truth, the \emph{Overall\ IoU} metric is the ratio between the intersection and the union of the two, which is calculated by dividing the total intersection area by the total union area. Both intersection area and union area are accumulated over all test samples. The \emph{Pre@X} measures the percentage of test examples that have \emph{IoU} score higher than the threshold $X$. In our experiments, $X \in \{0.5, 0.6, 0.7, 0.8, 0.9\}$.

\textbf{Implementation Details:}
Given an input image, we resize it to $480 \times 480$ and adopt Swin Transformer~\cite{liu2021swin} pre-trained on ImageNet-22K dataset~\cite{2009ImageNet} as our backbone, following previous works~\cite{yang2022lavt}. On all three benchmark datasets, we keep the maximum length of query expression as 20. The language model we use is a BERT~\cite{devlin2018bert} model with 12 layers, a hidden size of 768, and is initialized with official pre-trained weights. The number of inference stage $N$ is equal to 4. Our model is optimized with a binary cross-entropy loss, and we employ the AdamW optimizer~\cite{2017Decoupled} with a weight decay of $0.01$. We employ a learning rate schedule with an initial learning rate set to $5e^{-5}$ and a polynomial learning rate decay.
We use the batch size of 32 and train on 8 Tesla V100 with 16 GPU VRAM. During inference, we upsample the prediction results back to the original image size and use \emph{argmax} to select the index on the channel dimension of the score map, no other post-processing operations are required.
% In other words, we keep only the first 20 words of each expression. This is because most of the language expressions on the benchmark datasets are shorter than the predefined maximum length, which ensures the integrity of the input sentence in most cases.

\begin{figure*}
	\begin{center}
		\includegraphics[width=0.98\linewidth]{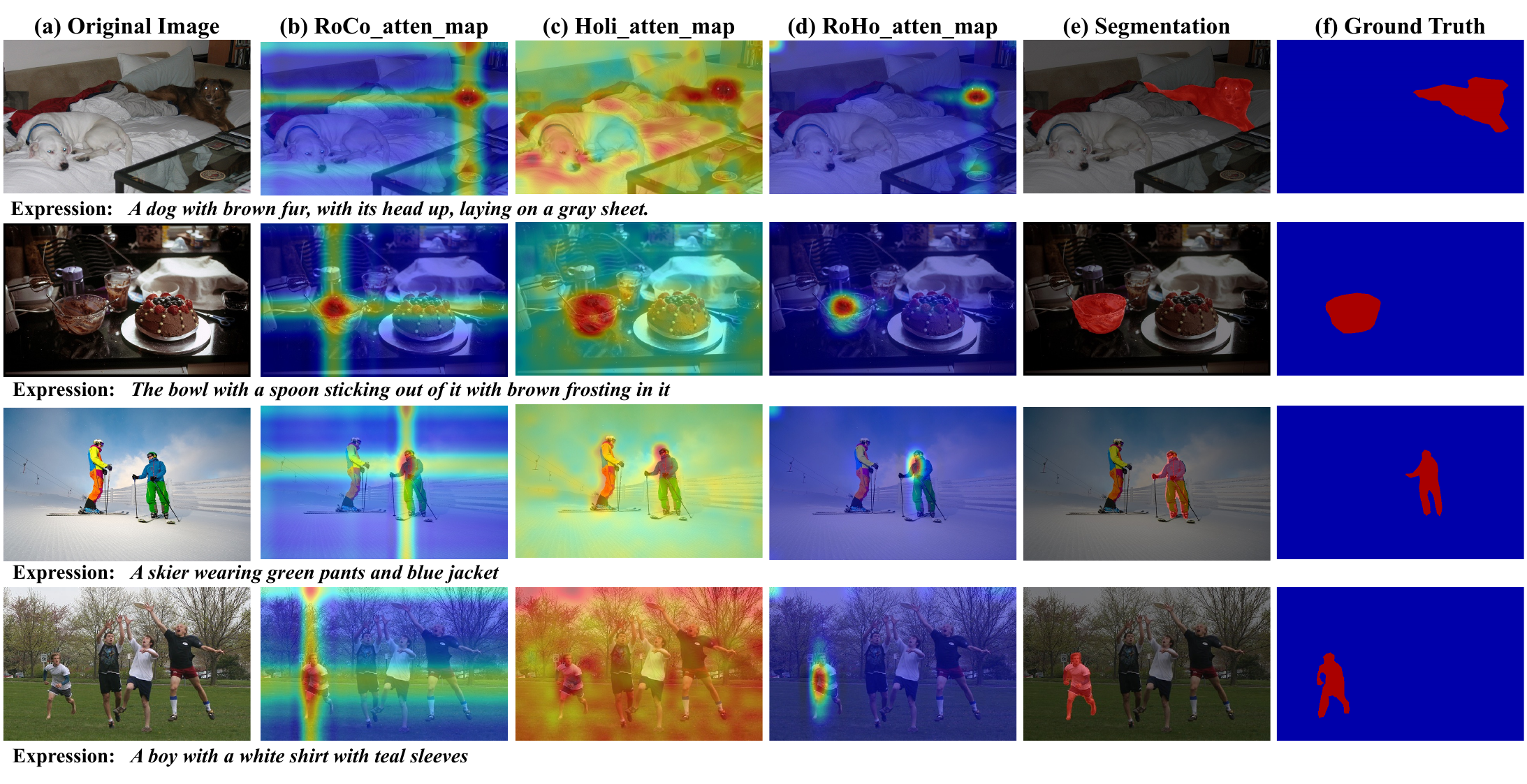}\\
		\vspace{-0.3cm}
		\caption{Visualization of the attention map of the Guided Attention layer. (a) Original image. (b) RoCo Attention map. (c) Holi Attention map. (d) RoHo Attention map. (e) Segmentation. (f) Ground-truth }\label{fig:6}
	\end{center}
\end{figure*}

\begin{figure*}
	\begin{center}
		\includegraphics[width=1.0\linewidth]{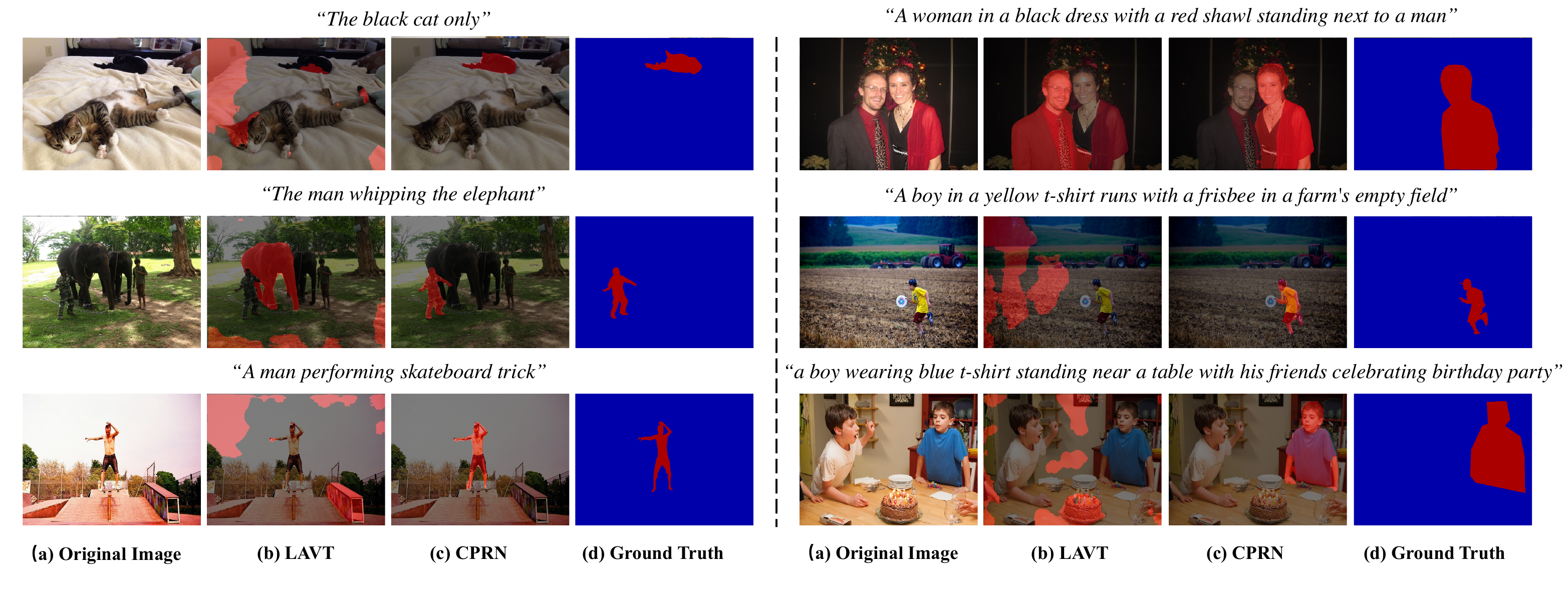}\\
		\vspace{-0.3cm}
		\caption{Visualization of comparisons between LAVT and CPRN segmentation results on the small-scale object and complex language set from the Gref-umd validation set. (a) Original image. (b) LAVT. (c) CPRN. (d) Ground-truth. }\label{fig:9} 
	\end{center}
\end{figure*}

\begin{figure*}
	\begin{center}
		\includegraphics[width=0.98\linewidth]{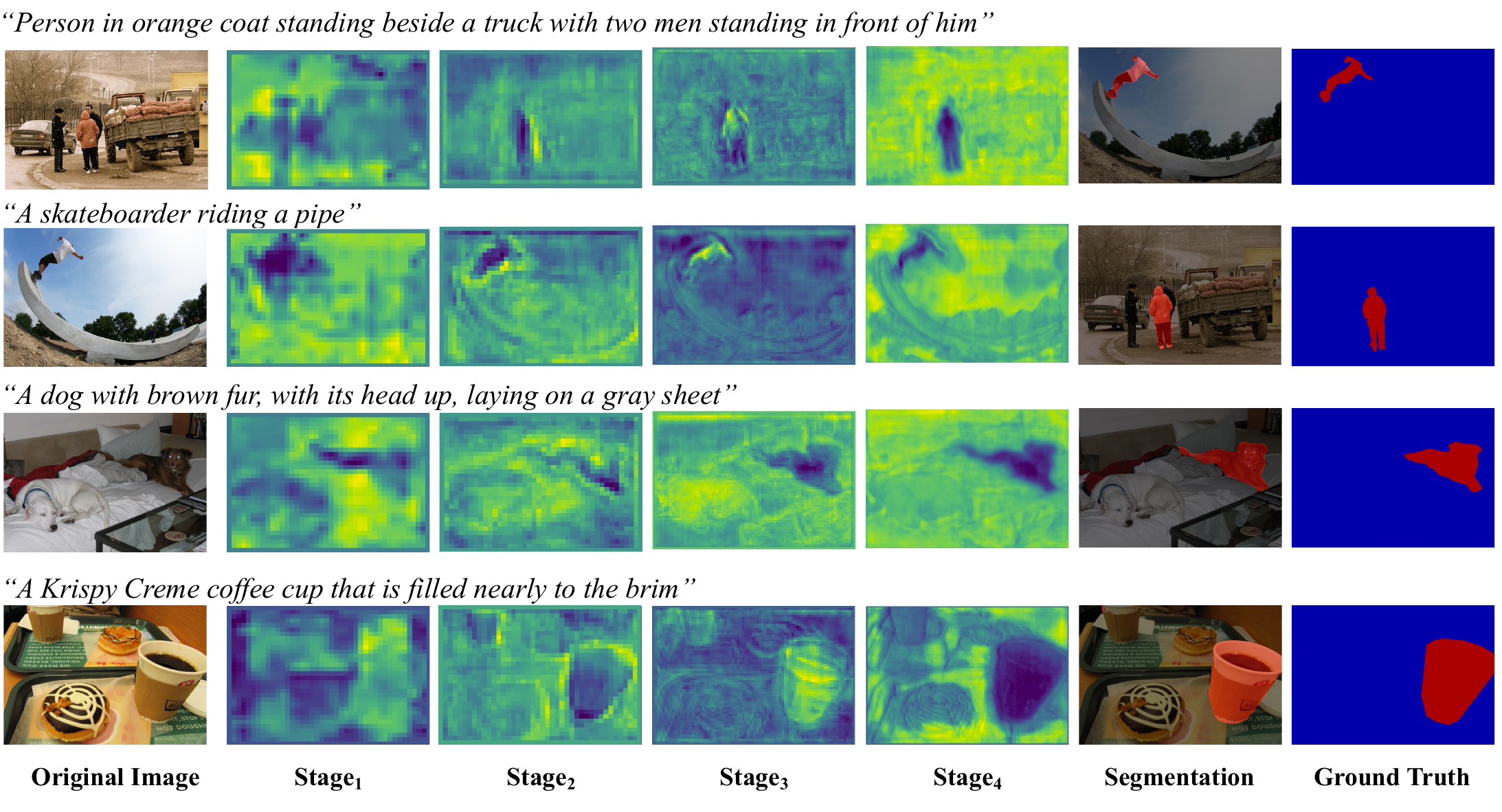}\\
		\vspace{-0.3cm}
		\caption{ Visualization of segmentation and feature map at different stages from the Gref-umd validation set. The left-most column shows the original image, and the right-most column illustrates the predicted mask and the ground truth mask.}\label{fig:7} 
	\end{center}
	\vspace{-0.3cm}
\end{figure*}

\begin{figure*}
	\begin{center}
		\includegraphics[width=0.98\linewidth]{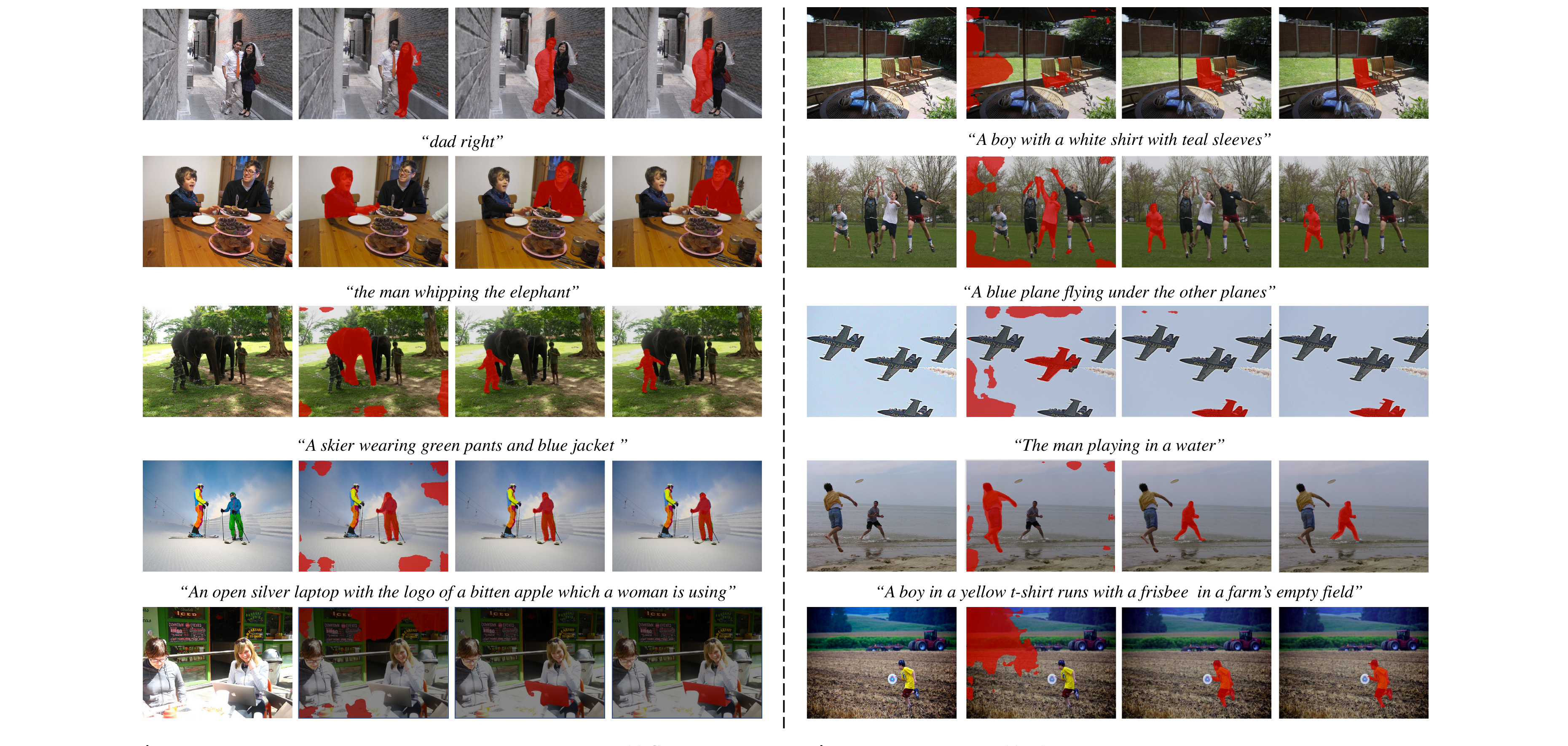}\\
		\vspace{-0.3cm}
		\caption{Visualization of comparisons between LSCM, LAVT and CPRN segmentation results from the RefCOCO+ testA set. (a) Original image. (b) LSCM. (c) LAVT. (d) CPRN. (e) Ground-truth. }\label{fig:8}
	\end{center}
	\vspace{-0.3cm}
\end{figure*}

\subsection{Quantitative Results}

\textbf{Comparison with State-of-the-arts:}
To demonstrate the superiority of our CPRN, we compare it with state-of-the-art methods, including RRN~\cite{2018RRN}, CSMA~\cite{2019Cross}, BRINet~\cite{2020Bi}, CMPC~\cite {huang2020CMPC}, LSCM~\cite{2020Linguistic}, EFN~\cite{2021Encoder}, BUSNet~\cite{yang2021bottom}, VLT~\cite{ding2021vision}, LTS~\cite{jing2021locate}, ReSTR~\cite {kim2022restr }, CRIS~\cite{wang2022cris} and LAVT~\cite{yang2022lavt} on three RIS benchmarks. The results of the comparison with other methods on three datasets using \emph{overall IoU} as metrics are displayed in Tab.~\ref{table:1}. 
It can be seen that our method consistently outperforms all previous methods on three benchmarks. 
In detail, CPRN yields an average improvement of 7.53\% over ReSTR on all three datasets. Similarly, it achieves a significant improvement in the range of 1.36\%-4.71\% compared to CRIS, which had previously achieved the best performance. 
In particular, our method also greatly outperforms LAVT using the same backbone, which fully illustrates the advantages of our proposed CPRN block. 
Specifically, on the refcoco+ dataset, our method exceeds LAVT by 1.44\% on val split, 1.06\% on testA split, and 0.74\% on testB split.
Since this dataset does not contain location information in referring expressions, it turns out that our CPRN block can enhance the ability to locate referents more than other methods that lack explicit modeling of positioning.
According to Tab.~\ref{table:1}, the gains are more obvious on other datasets than on the RefCOCO+ dataset, indicating that the ability of the CPRN block to locate language-responsive regions has a great impact on improving model performance.
More importantly, compared to the sota model, our network achieves 1.57\% and 2.16\% improvement on Gref-umd validation and test sets. 
Note that the referring expressions in the Gref dataset are generally longer, indicating that CPRN can better handle complex long sentences by positioning interactions between visual and linguistic features.
In the end, the above comparison with existing methods fully demonstrates the superiority of our method.

%-------------------------------------------------------------------------
\begin{table*}
\begin{center}
\caption{Comparison with LAVT on the Gref-umd validation set on a small dataset of small scale objects and complex languages.}
\label{table:3}
\setlength{\tabcolsep}{1.5mm}{
\renewcommand{\arraystretch}{1.2}
\begin{tabular}{l|c| c| c| c| c| c| c| c}
\hline
& {Methods} & \textbf{P@0.5} & \textbf{P@0.6} &  \textbf{P@0.7} & \textbf{P@0.8} &{\textbf{P@0.9}} & { \textbf{Overall IoU}} & \textbf{Mean IoU} \\ \hline
\hline
\multirow{2}{*}{small scale} & {LAVT}  & {43.96} & 36.24 & {28.32} & 18.81 & {2.57} & {28.54} & 41.63  \\
& {CPRN}  & {48.91} & 40.20 & {33.27} & 20.99  & {1.78} & {31.65} & 45.45  \\ 
\hline
\hline
\multirow{2}{*}{complex language} & {LAVT}  & {69.08} & 63.77 & {57.00} & 44.69 & {21.74} & {58.26} & 61.65  \\
& {CPRN}  & {72.46} & 70.29 & {64.73} & 52.66 & {25.12} & {61.88} & 64.77  \\ \hline
\end{tabular}}
\end{center}
\end{table*}

\begin{table}
\begin{center}
\caption{Experiments with five feature combination way of RoCo module on RefCOCO validation set.}
\label{table:4}
\setlength{\tabcolsep}{1mm}{
\renewcommand{\arraystretch}{1.2}
\begin{tabular}{c| c| c| c| c| c }
\hline
Methods & \textbf{P@0.5}  & \textbf{P@0.7} & \textbf{P@0.9} & \textbf{Overall IoU} & \textbf{Mean IoU} \\ 
\hline
\hline
  $f_1$  & 83.78  & 74.81  & 34.54 & 72.46 & 73.90 \\ 
  $f_2$  & 84.55  & 75.00  & 34.49 & 73.21 & 74.28 \\ 
  $f_3$  & 84.83  & 76.38  & 34.08 & 72.93 & 74.75 \\ 
  $f_4$  & 85.07  & 76.66  & 34.93 & 73.17 & 74.83 \\
\hline
\hline
CPRN & 85.09  & 76.54  & 35.24 & 73.42 & 75.00 \\ 
\hline
\end{tabular}}
\end{center}
\end{table}

\subsection{Ablation Studies}

    To investigate the relative contribution of each component in the proposed modules and the localization ability of the CPRN block, we conduct a series of ablation experiments on the RefCOCO dataset and evaluate it in the validation set, which is illustrated in Tab.~\ref{table:2}. In addition, we study the combination of row- and column-wise features in RoCo module, such as Tab.~\ref{table:4}. 
    Furthermore, to verify that our CPRN block is more effective on some non-salient objects with small scale and complex language expressions, we also conduct a plenty of experiments on the Gref-umd dataset and use \emph{overall IoU}, \emph{mean IoU} and \emph{Pre@X} as evaluation metrics, such  as Tab.~\ref{table:3}.

\textbf {Effective of RoCo module and Holi module. }
    To investigate the contribution of the Row-and-Column interactive (RoCo) module and Guided Holistic interactive (Holi) module to the overall model performance, we design four sets of ablation experiments. The baseline network is built with a simple holistic module (Holi* module) which only contains a cross-attention layer. We analyze: (1) Holi* module(Holi*), (2) RoCo module(RoCo), (3) RoCo module and Holi* module are combined in series(RoCo \& Holi*), (4) RoCo module and Holi* module are combined in parallel(RoCo $\Vert$ Holi*), (5) RoCo module and Holi module merged in two pathways(RoCo $\Vert$ Holi).
    The results in Tab.~\ref{table:2} show that since Holi* only uses a simple cross-attention mechanism, the segmentation performance of the model is not excellent. 
    Due to the RoCo module is only responsible for learning the modeling of positioning referents and lacks the perception of the global information of images, the result of the individual Roco module is poor. 
    In addition, we try the combination of the RoCo module and Holi* module in simple series and simple parallel, and both of the model performances do not improve greatly. 
    We also analyzed the reasons for the poor performance of serial combination.
    This is because the visual features are multi-modal interactively fused with language features in the RoCo module, and then sent into the Holi* module, the global information of the original visual features will be destroyed.
    For the parallel method, the model performance can already exceed using Holi* module alone. 
    Based on Holi module and RoCo module in parallel, our proposed CPRN has a better ability to localize referring entities than the scheme only using single Holi* module, scheme only using single Roco module, simple series scheme, and simple parallel scheme. 
    In order to further improve the performance, we have made some improvements, adding absolute postion embedding(ape) to the visual features and adding the FFN network layer after the combination of the RoCo module and the Holi module. Through experimental verification, these improvements will bring certain improvements to the model performance.

\textbf{Performance on small-scale objects and complex language expressions. }
    In order to demonstrate the effectiveness of our CPRN block, especially the positioning ability of small-scale non-salient objects, and the joint reasoning ability of complex language expressions, we constructed two small datasets on the Gref-umd dataset for further ablation studies. 
    In detail, we use thresholds of 0.03 and 18 as criteria for segmenting small-scale objects and complex language expressions, separately. 
    We consider data with the mask rate less than 0.03 as small-scale objects, and the proportion of this small dataset is 10.31\%.
    Besides, We consider data with language expression length longer than 18 after tokenizer as complex language queries, and this small dataset accounts for 8.46\%. 
    The result can be seen from Tab.~\ref{table:3} that CPRN outperforms the sub-optimal method LAVT by absolute advantage on those two reconstructed datasets with small-scale objects and complex languages, and is higher than the Overall IoU 3.11\% and 3.62\%. The visualization of these two small datasets also shows the advantage of our model in Fig.~\ref{fig:9}. The left column is the visualization of small-scale objects, and the right column is the visualization of complex language expressions.
    The visualization clearly illustrate that our model has great advantages, especially in solving the segmentation problem of small-scale objects and complex language expressions in the RIS task.

\textbf{The combination of row- and column-wise features. }
    In order to verify the combination of row- and column-wise muti-modal features, we designed four different fusion ways to generate $\mathbf{{v}_{hw}^{all}}$, which are the functions $f_1$, $f_2$, $f_3$ and $f_4$ as as follows:

\begin{equation}
    \begin{aligned}
          f_1 = \left( \mathbf{v}_{h}^{Att} \otimes \mathbf{v}_{w}^{Att} \right) + \textbf{V} 
	\label{eq:13}
	\end{aligned}
\end{equation}
\begin{equation}
    \begin{aligned}
        f_2 = \left( \mathbf{v}_{h}^{Att} \otimes \mathbf{v}_{w}^{Att} \right) * \textbf{V} 
	\label{eq:14}
	\end{aligned}
\end{equation}
\begin{equation}
    \begin{aligned} 
    % \begin{split}
		f_3 = {C} \left( {B} \left( \mathbf{v}_{h} + \mathbf{v}_{h}^{Att} \right), {B} \left( \mathbf{v}_{w} + \mathbf{v}_{w}^{Att} \right) \right)              
% 	\end{split}
	\end{aligned}
	\label{eq:15}
\end{equation}
\begin{equation}
    \begin{aligned}
% 	\begin{split}
         f_4 = {C} \left( {C} \left( {B}(\mathbf{v}_{h}), {B}(\mathbf{v}_{h}^{Att}) \right) , {C} \left( {B}(\mathbf{v}_{w}), {B}(\mathbf{v}_{w}^{Att}) \right) \right) 
% 	\end{split}
	\end{aligned}\label{eq:16}
\end{equation}
where $B$ denotes the Bilinear Interpolation and $C$ is a function implemented as concatenating in the channel dimension, and then connect a $1\times1$ convolution. Actually, the combination function we use in CPRN block is Eq.\ref{eq:5} and the experimental results in Tab.~\ref{table:4} also demonstrate that it is optimal way.

\subsection{Visualization}
    Fig.~\ref{fig:6} shows the visualization of attention maps in the Guided attention layer, which are RoCo Attention map (b), Holi Attention map (c), and RoCo and Holi Attention map(d). In order to better highlight the positioning effect of our CPRN block, the visualization of the RoCo attention map is the effect of superimposing the row and column features together. As can be seen from the visual attention map, our RoCo attention map can assist Holi attention in distinguishing confusing instance objects and guide the Holi module to segment the correct reference object. 
    Fig.~\ref{fig:7} is a visualization of the features at different stages, we can observe that the feature maps of each stage in CPRN (ie $\rm Stage_4$, $\rm Stage_3$, $\rm Stage_2$, $\rm Stage_1$) can accurately locate the semantic concepts referred by natural language expression. Fig.~\ref{fig:8} shows the visualization results of LSCM (Fig.~\ref{fig:8}(b)), LAVT (Fig.~\ref{fig:8}(c)), CPRN (Fig.~\ref{fig:8}(d)) and ground-truth (Fig.~\ref{fig:8}(e)) on RefCOCO+ testA set. The referring expressions in the RefCOCO+ dataset do not include words representing spatial or positional information, which places higher demands on the ability to understand the appearance of objects. A comparison of visualization results shows that our proposed CPRN block can positioning referring entities effectively even in the absence of explicit location information. Furthermore, we are able to reason about complex textual information to obtain the final segmented referents.

    Taking the first case as an example, given the query expression "back to us white", CPRN can obtain the location information of the instance referent. Moreover, for the language expression "man on floor", our method can also accurately locate the target.
    Both LSCM and LAVT cannot accurately localize language-responsive region relying solely on global information of visual image.
    Furthermore, it can be seen that our CPRN can accurately segment referring objects even in language representations of different lengths and complex scenes, such as "black coat looking at suitcase black hat", LSCM model mis-segments multiple objects, LAVT model is determine difficultly whether the segmentation goal is "black coat" or "suitcase black hat", but CPRN can distinguish segmentation objects, and rows 4 and 5 also show similar results. 
    
    From the overall visualization results, it can be seen that CPRN can accurately positioning entities without bringing too much redundant mask information, which is crucial for improving the performance of referring
    image segmentation task.

\section{Conclusion}

In this work, we propose a collaborative positioning reasoning network for referring image segmentation task, which can efficiently locate the referring entities with detailed edges, even small-scale objects or incomprehensible natural languages Express. Under the architecture of multi-semantic inference network, we adopt RoCo module and Holi module in parallel for each semantic stage. After dividing the overall visual feature map into horizontal direction map and vertical direction map, RoCo Module fuses them with texture information respectively, and the position of referring objects can be more accurate. 
The Holi module preserves the overall feature map to ensure the integrity of the global information, while the RoCo module guides the Holi module through a global guided pathway to generate correct segmentation results. 
The proposed method achieves state-of-the-art performance on three challenge benchmarks.

% In this work, we propose a collaborative positioning and joint reasoning scheme for the referring image segmentation task, which can effectively locate the referent, even of small scale, with detailed edges. 
% Under the architecture of multiple semantic inference network, we adopt both RoCo module and Holi module in parallel for each semantic stage.
% After dividing the holistic visual feature map into horizontal and vertical directional maps, the RoCo Module fuses them with texture information separately.
% The location of referents can be more accurate through their collaborative positioning. 
% The Holi module keeps the holistic feature map to ensure the integrity of global information. 
% The proposed method achieves state-of-the-art performance on three challenge benchmarks.

% \section{References Section}
% \bibliography{IEEEabrv,../bib/paper}
%\bibliography{egbib}
\bibliographystyle{IEEEtran}

\end{document}